# AN ADAPTIVE NEURO-FUZZY INFERENCE SYSTEM MODELING FOR GRID-ADAPTIVE INTERPOLATION OVER DEPTH IMAGES

Arbaaz Singh

Department of Computer science & Engineering,
Indian Institute of Technology, Ropar, Punjab, India
arbaazs@iitrpr.ac.in

## ABSTRACT

*A suitable interpolation method is essential to keep the noise level minimum along with the time-delay. In recent years, many different interpolation filters have been developed for instance H.264-6 tap filter, and AVS- 4 tap filter. The present work uses Adaptive Neuro-Fuzzy Inference System (ANFIS) technique to model and investigate the effects of a four-tap low-pass tap filter (Grid-adaptive filter) on a hole-filled depth image. The work demonstrates the general form of uniform interpolations for both integer and sub-pixel locations in terms of the sampling interval and filter length of depth-images via diverse finite impulse response filtering schemes. The demonstrated model combined modelling function of fuzzy inference with the learning ability of artificial neural network.*

## 1. INTRODUCTION

In 3-D computer graphics, a depth map is an image or image channel that contains information relating to the distance of the surfaces of scene objects from a viewpoint [1]. Once the original image and depth image is given, 3-D can be synthesized by mapping pixel coordinates one by one according to its depth value. It is the next emerging revolution after the high definition video and is the key technology in advanced three dimensional television systems (3-D TV) and free-view television systems [2-4]. A new member of 3-D sensor family, kinect has drawn great attention of researchers in the field of 3-D computer vision for its advantage of consumer price and real time nature. Based on a structured light technique, Kinect is able to generate depth and colour images at a speed of about 30 fps [5]. However, limited by depth measuring principle and object surface properties, the depth image captured by the Kinect contains missing data as well as noise. These areas of missing data are known as holes. Holes appear due to sharp horizontal changes in depth image, thus the location and size of holes differ from frame to frame. Several attempts are made to remove the noise and filling of holes with the correct data to make it suitable for different applications by means of bilateral and median-filters [6]. Apart from hole-filling, image interpolation as well occurs in all digital pictures at several stages [7]. Interpolation is the process of determining the values of a function at positions lying between its samples. It achieves this process by fitting a continuous function through the discrete input samples. This permits input values to be evaluated at arbitrary positions excluding those defined at the sample points. Interpolation is required to produce a larger image than the one captured and finds an imperative consign in transmission of 3-D images. 3-D images have been used in robotic guidance, product profiling and object tracking, in battle preparation, medical diagnosis and many more [8]. These all applications require both hole-filling and interpolation to provide a preferred output. A suitable interpolation method is essential to keep the noise level minimum along with the time-delay. The standard/conventional procedure to interpolate a depth image is to first fill the holes and then apply the interpolation filter that leads to low $PSNR$ and a great time-complexity. In recent years, many different interpolation filters have been developed for instance H.264-6 Tap filter, AVS- 4 tap filter and so on [9-10]. The standard/conventional procedure to interpolate a depth image is to first fill the holes and then apply the interpolation filter that leads to low $PSNR$ and a great time-complexity. Recently, a texture-adaptive hole-filling algorithm is proposed for post-processing of rendered image on 3D video to save the computational cost [11]. The algorithm first determines the type of holes, and then fills the missed pixels in raster-order depending upon hole types and texture gradient of neighbours with simple data operation, which benefits for fast processing. Further, the quality of virtual view images from rendering is demonstrated by establishing a connection of pixel coordinate warping between reference image and virtual image in the rendering process, to make a quick decision on the position of hole-region and its edge [12]. Subsequently, by outward expanding the edge of holes, warping error pixels are covered. Then, hole-filling is through using use mean filter and the image restoration method after eliminating the false contour by image synthesis. Nonlinear filters called Spline Adaptive Filters (SAFs), implementing the linear part of the Wiener architecture with an IIR filter instead of an FIR one are also come into picture to improve the PSNR and computational delay [13]. Some analytical models have been studied using Fuzzy models in literature to model and investigate the impact of Grid adaptive filters over depth images. An Adaptive Neuro-Fuzzy Inference System (ANFIS) based models of grid adaptive filters are modelled by employing optimized membership functions to develop Fuzzy Inference System (FIS). This paper investigates a four-tap Grid-adaptive filter on a hole-filled depth image that provides uniform interpolations for both integer and sub-pixel locations in terms of the sampling interval and filter length. Furthermore, an ANFIS model is designed and analyzed to support our experimental results for different finite impulse response filters to recommend the best one.

## 2. EXPERIMENTAL ANALYSIS

The depth image $I$ is a 2D grid of $K_V \times K_H$ pixels. The pixels denote the distance of the objects in real scene according to its image co-ordinates [4]. The holes are areas in the image that have invalid (missing) data. Each pixel has 8 neighbouring pixels that share a face or a vertex with the centre pixel. Our goal is to go through the image and fill up the holes in the image. Let each pixel in the image be denoted as $I(v, h)$, where $v, h$ can be any integer value between $0$ and $K_V$, $K_H$ respectively. If $I(v, h) = 0$, then that pixel is a part of a hole. A $3 \times 3$ gaussian weighted averaging filter is used to fill the holes. The filter takes the weighted average of the depth values of its neighbours and replaces the hole with the obtained averaged value. The weights are decided such that the neighbour pixels, that are holes themselves, are ignored while the non-zero valued pixels are used to find the new value of the hole-pixel. Let the vertical and horizontal distances between the two nearest known pixels be $T_v$ and $T_h$, respectively. Subsequently, the aim is to insert and interpolate $N_v$ and $N_h$ pixels within the intervals of $T_v$ and $T_h$ respectively. So, after the interpolation, we will have a sum of $[K_v(1+N_v) - N_v] \times [K_h(1+N_h) - N_h]$ pixels and the sampling intervals for vertical and horizontal directions will be changed to $D_v = T_v/(1+N_v)$ and $D_h = T_h/(1+N_h)$ respectively. For example, Figure 1 shows the case of $N_v = N_h = 1$ with $D_v = T_v/2$ and $D_h = T_h/2$.

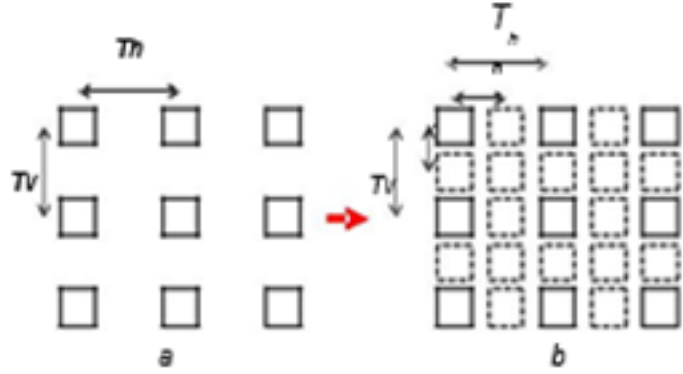

Fig. 1: Uniformly spaced interpolation (a) before interpolation with known (bold square) pixels, (b) After Interpolation for missing (dotted square) pixels

The proposed-4 tap filter is obtained using the Langrage interpolation, which is used to generalise the linear interpolation by approximating the *sinc* function [10]. For any sampling grid layout and scale, the filter coefficients can be calculated by fitting (3) to the grid. The $PSNR$ is measured for different filters such as the linear averaging filter, H.264-6 Tap filter with coefficients *(1,-5, 20, 20, -5, 1 ) / 32*, the AVS- 4 Tap filter with coefficients *( -1, 5, 5,-1 ) / 8* and the proposed-4 tap grid adaptive filter with coefficients *( -1, 9, 9, -1 ) / 16* to recommend the best one. All test images are taken from the Middlebury Database *(2006)*. For evaluating the performance of the simulated filters, all images are expanded at the same sampling layout and scale of the filters (i.e. doubling the number of rows and columns with $T=2, N=1, M=2$, and $D=1$). For computation, the images are reduced to half before interpolation so that the size of image remains same after hole-filling and interpolation. The PSNR is calculated between a perfect image and its noisy approximation and can be easily defined via mean squared error ($MSE$). For a given noise-free $m \times n$ monochrome image $'I'$ and its noisy approximation $'K'$, $MSE$ is defined as:

$$MSE = \frac{1}{mn}\sum_{i=0}^{m-1}\sum_{j=0}^{n-1}(I(i,j)-K(i,j))^2 \qquad \ldots (1)$$

and PSNR is measured as

$$PSNR = 10\log_{10}\left(\frac{I^2_{MAX}}{MSE}\right) = PSNR = 20\log_{10}\left(\frac{I_{MAX}}{MSE}\right) \qquad \ldots (2)$$

As shown in the Table 1, although the filter length is shorter than H.264-6, the proposed 4-Tap filter yields the highest average PSNR. The obtained results show that the proposed-4 filter with the coefficients *(-1, 9, 9, -1) / 16* has almost equal influence to the H.264 filter in terms of $PSNR$ evaluation. Consequently, it is suggested to recommend H.264-6 filter to produce high quality images. Otherwise, the proposed-4 filter is a better option that provides high quality images along with minimal time processing delays as it is used four level tapping.

**Table 1. Measured PSNR using different Filters**

| Depth Images | Measured PSNR | | | |
|---|---|---|---|---|
| | Linear Average Filter | AVS- 4 Tap Filter | H.264- 6 Tap Filter | Grid Adaptive-4 Tap Filter |
| Aloe | 39.9108 | 40.0318 | 40.0684 | 40.0846 |
| Baby 2 | 47.109 | 47.2672 | 47.2149 | 47.2845 |
| Baby 3 | 47.1809 | 47.4044 | 47.2218 | 47.4055 |
| Bowling 2 | 44.2 | 44.4352 | 44.3577 | 44.4295 |
| Cloth | 47.6444 | 47.7766 | 47.7397 | 47.8162 |
| Cloth 3 | 49.6144 | 49.7569 | 49.6258 | 49.7836 |

## 3. Fuzzy Analysis

In this section, the work is optimized via ANFIS by tailoring the membership functions for optimization of FIS of diverse interpolation filtering schemes for instance Linear Average Filter AVS- 4, Tap Filter, H.264- 6 Tap Filter and Grid Adaptive-4 Tap Filter. The lower two layer's parameters are common to all depth images taken into account. In a fuzzy system, certain inputs are given to the rules, depending upon the system under consideration. The output given by the system is then utilized to reach the final decision. The whole process is clear from a typical fuzzy inference system shown in Fig. 2.

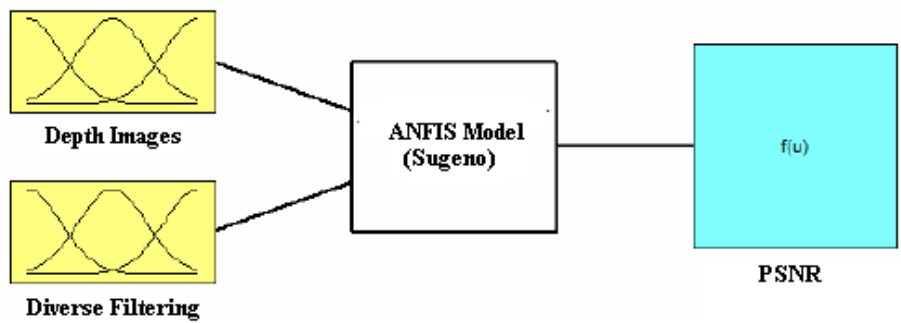

Fig. 2: ANFIS Model for diverse Filtering schemes

A fuzzy inference system is developed by using Matlab$^{TM}$ software for five variables such as size of depth images, Linear Average Filter AVS- 4, Tap Filter, H.264- 6 Tap Filter and Grid Adaptive-4 Tap Filter. For each of these variables, the universe of discourse and the degree of membership function are defined as shown in Fig 3-11. Gaussian membership functions are used in the proposed fuzzy inference system to define the shape of both input and output variables. We have evaluated one output variable such as PSNR depending upon the five input variables as shown in Fig 3-6. The output variables illustrate the possible different outcomes for recommending the optimum interpolation filtering scheme for the depth images. The fuzzy logic is based on the identification of the fuzzy-set that represents the possible values of the variables. Fuzzy model described in this paper is a MISO system with five input parameters. Possible universe of discourse for the input parameters is given below:

**Inputs parameter:** Depth Images; Diverse Filtering Schemes = Linear Average Filter AVS- 4 Tap Filter, H.264- 6 Tap Filter and Grid Adaptive-4 Tap Filter

**Output parameters:** PSNR

ANFIS is a five layered network i.e. Fuzzification, Rules, Normalization, De-fuzzification and output layers, in which each layer performs a particular task. Each layer consists of number of nodes performing the similar functions. The forebear parts of fuzzy rules are represented by nodes in first layer and every node is an adaptive node. Fuzzification layer determines the degree of membership functions of inputs i.e. Transmission time and diverse routing algorithms. Layer 2 is the rule layer having fixed nodes. The output of this layer represents the firing strength of each rule. Firing strength is obtained by fuzzy logic operator "intersection" which gives the product of the input membership grades. Layer 3, normalization layer also consists of fixed nodes. Each node in this layer receives inputs from all nodes in the rule layer, and calculates the normalized firing strength of a given rule. Layer 4 is the de-fuzzification layer having adaptive nodes. A de-fuzzification node calculates the weighted consequent value of a given rule and is simply the product of the normalized firing strength and the first order polynomial (for a first-order Sugeno model). Layer 5, output layer consists of a single fixed node and the output of node is calculated as the summation of all incoming signals. In our work, the impact of diverse interpolation filtering is investigated for PSNR as output variable over depth images as input variable. The tailored membership functions are depicted in Fig (3-11) using ANFIS.

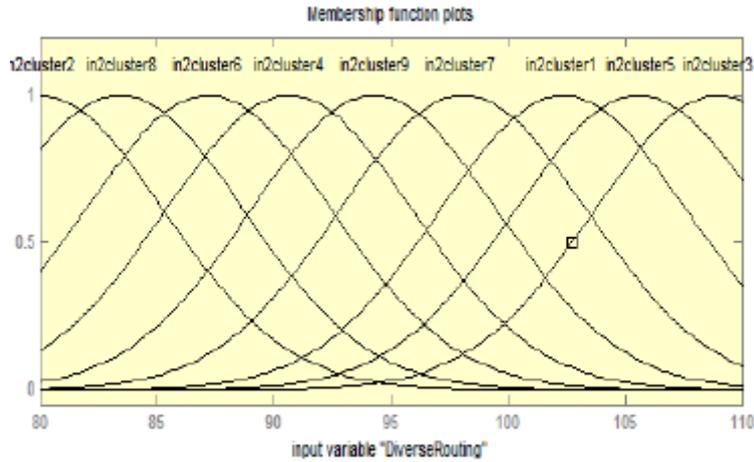

Fig 3: Tailored Membership Functions for diverse Filtering

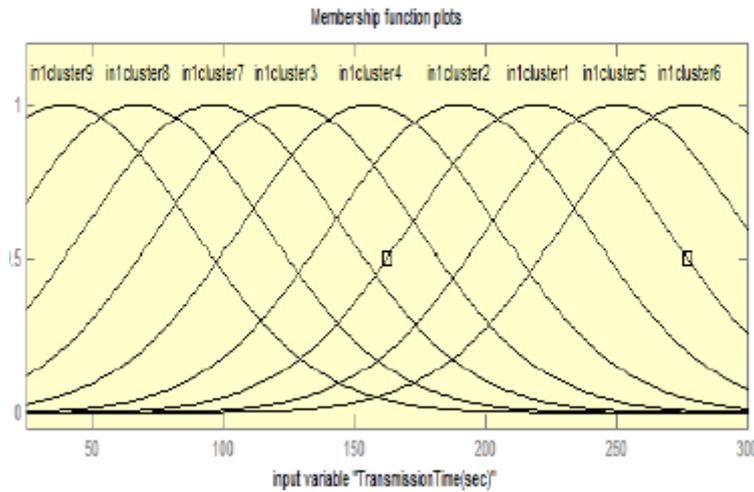

Fig 4: Tailored Membership Functions for diverse Sizes of Depth Images

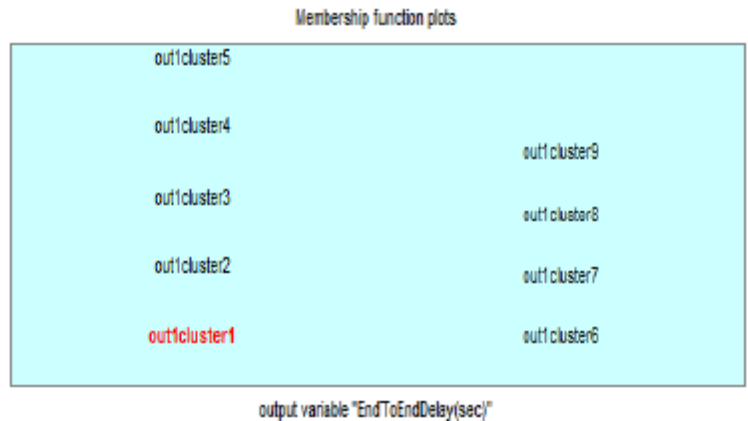

Fig 5: Tailored Membership Functions for PSNR

After designing the membership functions for different input and output variables, the next step is to define the rules that will govern these variables. After going through different iterations and removal of non-contributing rules, for End to End Delay vector, having five inputs, the rules are optimized to nine rules as shown in Table 7. Fig. 12 indicates rule viewers that show the values of the various defined inputs to the model and the corresponding computed output. The calculated Average Testing Error comes to be 0.059 only. Following a similar approach, rules for vectors Throughput and Retransmission attempt were defined as shown in Table 8-9 and Fig. 13- 14. Each component of each vector can be considered as a function of all components of the preceding vector. The beauty of such systems is their ability to provide the same output that a decision maker would provide in any given situation, from a set of given inputs. To further enhance the reliability of the proposed system, a series of simulations was carried out by varying one input at a time or many inputs simultaneously. After carrying out the above-mentioned simulations, several evaluations were made in the final stage of the proposed fuzzy inference systems. For each input, a specific value is given to define the intermediate vectors, using the fuzzy rules. It is ensured that the values of each input are increased and decreased from almost its minimum to its maximum. This makes the working of the proposed system more understandable to get the final output for a wide range of values. The various simulations are illustrated, by using different input values. These simulations are demonstrated not only the working of MANET, but also, the variations in the values obtained for intermediate vectors that helps in drawing the final recommendation. ANFIS incorporates fuzzy "if–then" rules involving premise and consequent parts of Sugeno type FIS. For the first order, rule set with nine fuzzy "if–then" rules for PSNR via diverse filtering schemes is depicted in Table 2. The Grid Adaptive-4 Tap Filter is depicted as 1, Linear Average Filter as 2, AVS- 4 Tap Filter as 3 and H.264- 6 Tap Filter as 4 in the IF-THEN Rules.

**Table 2: IF-THEN Rules for evaluated PSNR of diverse Depth images**

1. If (Size of Depth Image is in1cluster1) and (Filter is in2cluster1) then (PSNR is out1cluster1) (1)
2. If (Size of Depth Image is in1cluster2) and (Filter is in 2cluster2) then (PSNR is out1cluster2) (1)
3. If (Size of Depth Image is in1cluster3) and (Filter is in2cluster3) then (PSNR is out1cluster3) (1)
4. If (Size of Depth Image is in1cluster4) and (Filter is in2cluster4) then (PSNR is out1cluster4) (1)
5. If (Size of Depth Image is in1cluster5) and (Filter is in2cluster5) then (PSNR is out1cluster5) (1)
6. If (Size of Depth Image is in1cluster6) and (Filter is in2cluster6) then (PSNR is out1cluster6) (1)
7. If (Size of Depth Image is in1cluster7) and (Filter is in2cluster7) then (PSNR is out1cluster7) (1)
8. If (Size of Depth Image is in1cluster8) and (Filter is in2cluster8) then (PSNR is out1cluster8) (1)
9. If (Size of Depth Image is in1cluster9) and (Filter is in2cluster9) then (PSNR is out1cluster9) (1)

For the first order, the control surface of fuzzy model with nine fuzzy "if–then" rules for PSNR via diverse filtering schemes is depicted in Fig. 6 and reveals out that Grid adaptive filters provides better PSNR as compare to all other tested filters.

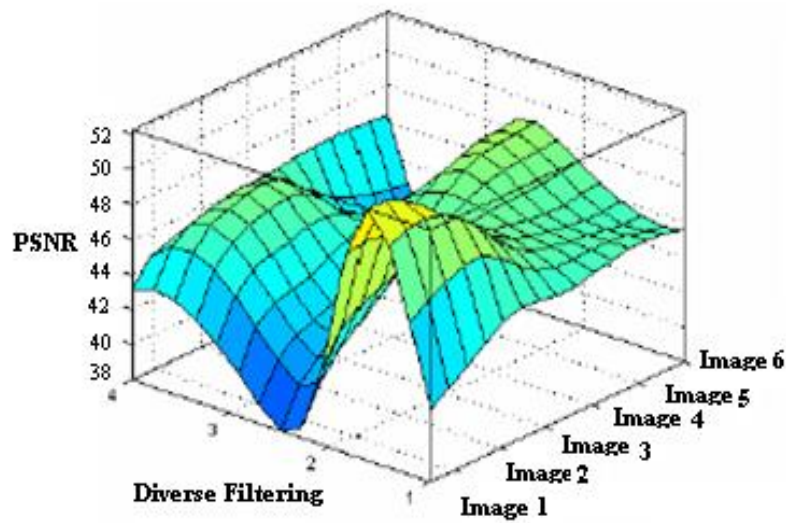

Fig 6: Control surface of fuzzy model showing inter-dependency of diverse Interpolation filtering over PSNR

## 3. CONCLUSIONS

To demonstrate the power of grid adaptive filter, the PSNR is measured for different filters to determine the best one as our analysis has shown. In our experiments, it has been shown that the grid adaptive four-tap filter yields the highest average *PSNR* values (almost same as that of the six-tap filter). The results are also supported by ANFIS analsys reported in section 3. Accordingly, it is suggested to use the proposed integrated Grid adaptive filter for the enhancement of depth images.

## ACKNOWLEDGEMENT

I thank to Professor Chee Sun Won, Department of Electronics and Electrical Engineering, Dongguk University, Seoul, South Korea for his kind and valuable guidance for carrying out this work. I also thank to the Department of Electronics and Electrical Engineering, Dongguk University, Seoul, South Korea for providing lab facility and technical support.

.

## Author's Biography:

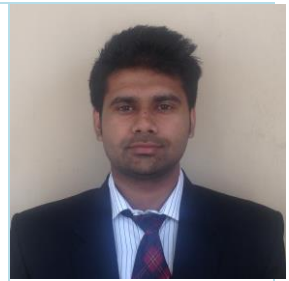

I, Arbaaz Singh am currently studying in the field of Image Processing in the department of Computer Science & Engineering at Indian Institute of Technology, Ropar, Punjab, India. I worked as a research Intern at Dongguk University under the kind guidance of Professor Chee Sun Won in the field of Depth Images. My areas of interest are Depth image processing and designing issues of Interpolation filters.